\documentclass[11pt]{article}

\usepackage[utf8]{inputenc}
\usepackage[T1]{fontenc}
\usepackage{amsmath,amssymb}
\usepackage{algorithm}
\usepackage{algpseudocode}
\usepackage{float}

\usepackage{booktabs}
\usepackage{graphicx}
\usepackage{comment}
\usepackage{listings}
\usepackage{xcolor}
\usepackage[margin=1in]{geometry}
\usepackage{caption}
\usepackage{hyperref}
\usepackage{colortbl} 
\usepackage{longtable} 
\definecolor{gaphi}{RGB}{248,233,228} 
\newcommand{\gapc}[1]{\cellcolor{gaphi}#1}
\newcommand{\ttsplit}[2]{\begin{tabular}[t]{@{}l@{}}\texttt{#1}\\\texttt{#2}\end{tabular}}

\title{\textbf{Self-Modifying Lean Proof Agents with\\
Verifier-Grounded Benchmark Coevolution}}
\author{
  Yuqing Li \\ University of Pittsburgh \\ \texttt{yul658@pitt.edu}
  \and
  Zeguan Wu \\ University of Pittsburgh \\ \texttt{zew79@pitt.edu}
  \and
  Yu Gan \\ University of Pittsburgh \\ \texttt{yug130@pitt.edu}
  \and
  Junyu Liu\thanks{Corresponding author.} \\ University of Pittsburgh \\
  \texttt{junyuliucaltech@gmail.com}
}
\date{}

\begin{document}
\maketitle

\begin{abstract}
Designing effective Lean proof agents is a central challenge in formal
mathematical reasoning. While one line of work builds stronger Lean-oriented
provers, recent perspectives emphasize that theorem-proving agents also need
better workflows around Lean: how they decompose proof obligations, use tools
and compiler feedback, diagnose failed attempts, repair proofs, and maintain
structured proof context. Motivated by code-level self-evolving agents, we
study whether such workflows can be evolved rather than hand-designed. We
present a self-evolving
Lean proof agent in which a small \emph{fixed, trusted} runtime wraps a fully
\emph{mutable} workspace --- the proof workflow, prompts, and tools. Unlike most
self-evolving agent systems, which optimize against an externally supplied, fixed
benchmark, our system coevolves the agent and its benchmark. Between
generations, the highest-scoring agent --- the \emph{champion} --- revises the active
task distribution through a \emph{mastery-throttled} curriculum update,
introducing harder proof obligations only after the current level has been
mastered; a \emph{single-anchor recalibration} then re-runs the champion on the
updated benchmark so that scores remain comparable as difficulty rises.
All evolution stays inside a Lean-grounded verification loop: however the agent
rewrites itself, a success counts only when its behavior yields Lean-verified
proofs under a trusted snapshot. Each solve attempt must also emit a
machine-readable, Lean-grounded \emph{proof context} whose representation is
free to evolve but whose groundedness is enforced. We run both the coevolving
trajectory and a fixed-benchmark baseline for $15$ active generations. Because
coevolving raw scores are measured under changing benchmark difficulty, we
compare runs on a held-out miniF2F test split, evaluating selected agents from
the coevolving trajectory and the two highest-scoring agents from the
fixed-benchmark baseline. On this split, the best evaluated coevolving agent
reaches a $45.1\%$ held-out solve rate, compared with $12.7\%$ for the seed and
$32.0\%$ for the best fixed-benchmark baseline agent. These results show that verifier-grounded
self-evolution can improve Lean proof workflows under a coevolving benchmark.
\end{abstract}

\section{Introduction}
\begin{figure}[t]
	\centering
	\includegraphics[width=\linewidth]{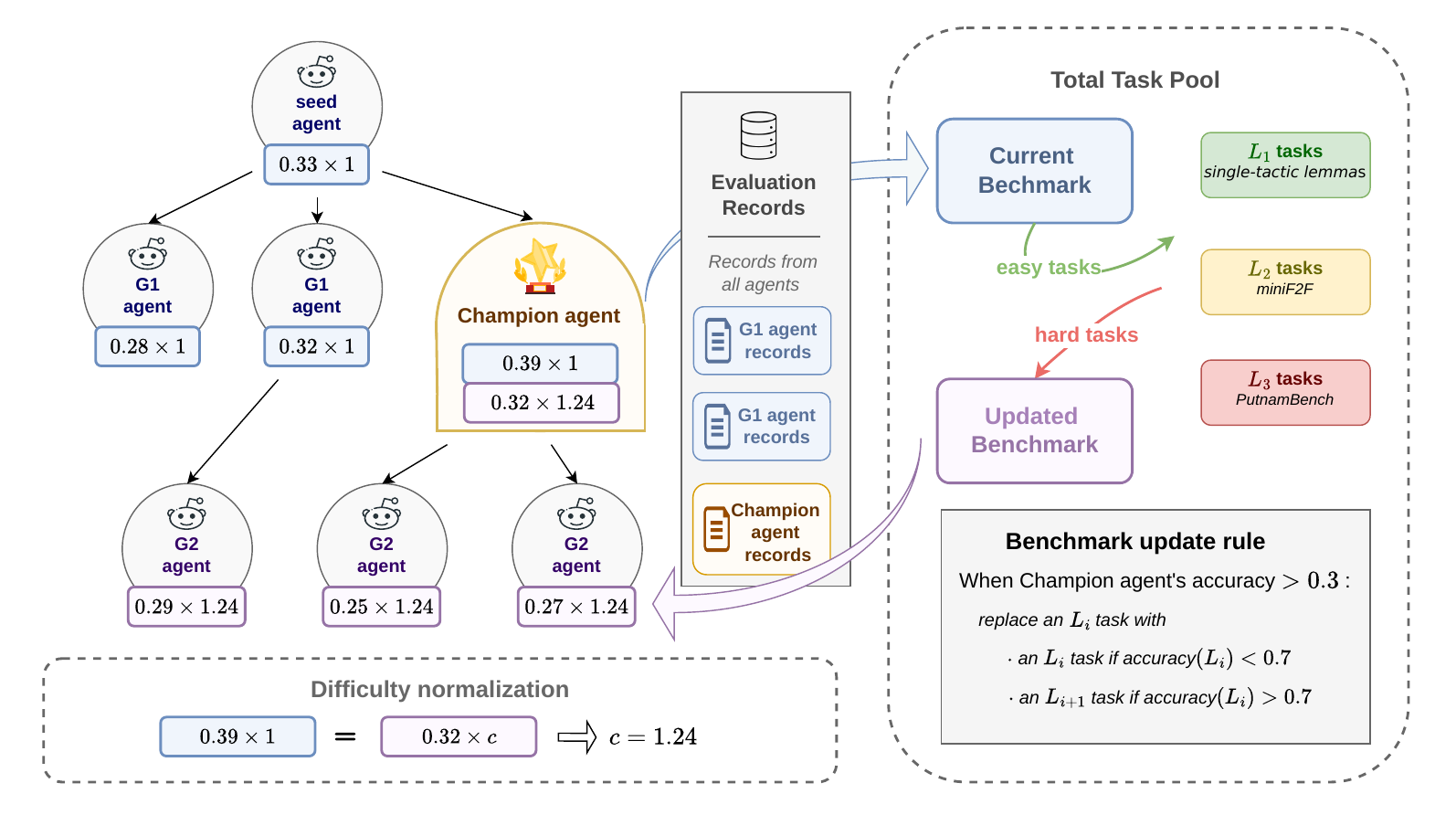}
	\caption{\textbf{Champion-driven agent--benchmark coevolution.}
		A \emph{seed agent} produces first-generation (\textbf{G1}) children that
		stream into a shared archive as they are evaluated, so a strong early child
		can already parent later siblings in the same generation; the strongest
		child becomes the \emph{champion}, seeds the next generation (\textbf{G2}),
		and also drives the benchmark update. Evaluation records from
		all agents are collected. The benchmark is drawn from a candidate problem
		pool stratified by difficulty level ($L_1$: single-tactic lemmas; $L_2$:
		miniF2F valid split, disjoint from the held-out test set used only
		for generalization; $L_3$: PutnamBench). When the champion clears the
		update threshold, mastered tasks are retired and replaced either laterally by
		new tasks from the same level, or by harder tasks from the next level
		once that level has been mastered. The updated benchmark is then used to
		evaluate the next generation. To keep scores comparable as the benchmark
		becomes harder, the champion is re-run on the updated benchmark as a
		single-anchor recalibration: equating its normalized scores, e.g.\
		$0.39\times 1 = 0.32\times c$, yields the new difficulty coefficient
		$c=1.24$ (illustrative values).}
	\label{fig:overview}
\end{figure}

\paragraph{Lean proof agents and the workflow bottleneck.}
Designing effective agentic frameworks for Lean-based theorem proving has
become an increasingly central problem in formal
mathematical reasoning~\cite{alphaproof,deepseekproverv2,goedelproverv2,leap,goedelarchitect}.
One line of work aims to build stronger Lean-oriented provers: specialized
models, tactics, or search procedures that more directly synthesize valid Lean
proofs~\cite{alphaproof,deepseekproverv2,goedelproverv2}. A complementary,
workflow-centered line --- exemplified by LEAP~\cite{leap} and
Goedel-Architect~\cite{goedelarchitect} --- instead emphasizes that the
bottleneck is not only prover strength, but also how an agent interacts with
the Lean proof environment: how it decomposes proof obligations, invokes
tools, uses compiler feedback, diagnoses failed attempts, repairs proofs, and
maintains useful proof context. Both rely on hand-designed proof workflows built
around intermediate Lean lemmas or AND/OR-style decomposition, and
report strong results (Goedel-Architect reaches $99.2\%$ on miniF2F-test),
showing that the proof workflow itself is a major determinant of performance.

This work follows the workflow-centered view. We do not aim to make Lean
itself a stronger verifier. Instead, Lean serves as a fixed, trusted substrate
for checking correctness, while the object of improvement is the agentic proof
workflow around it. Our motivating question is therefore: if a Lean proof agent
is not given a carefully hand-designed workflow, what proof workflow can it
evolve for itself?

\paragraph{From DGM to Hyperagents.}
We approach this question through recent work on code-level self-evolving
agents. The Darwin G\"odel Machine (DGM)~\cite{dgm} showed that an agent can
improve by rewriting its own code and empirically validating the resulting
variants on downstream tasks. This makes DGM a natural starting point for
self-improvement, but its original setting is closely aligned with coding: the
agent modifies code, and its improvement is measured on coding benchmarks.

Hyperagents~\cite{hyperagent} extend this line by introducing a
self-referential architecture that integrates two roles into a single editable
program: a \emph{task agent}, which solves the target task, and a \emph{meta
	agent}, which modifies both itself and the task agent. Because the meta-level
modification procedure is itself editable, the system can improve not only
task-solving behavior but also the mechanism that generates future
improvements. This distinction matters for formal mathematics: it allows a self-evolving
framework to target Lean theorem proving, where the object of improvement is
the mathematical proof workflow itself rather than the agent's general coding
ability.

\paragraph{From self-evolving agents to coevolving benchmarks.}
Although coevolving environments have been studied in open-ended systems
such as POET and MCC~\cite{poet,mcc}, many recent self-evolving agent systems
still evolve the agent while keeping the external environment
fixed~\cite{dgm,hyperagent,mae,godelagent,adas,aflow}. This is especially problematic for Lean theorem
proving, where benchmark difficulty determines whether evolution receives an
informative selection signal. In our experiments, the seed proof agent solves only
$12.7\%$ of the held-out miniF2F test split overall, and solves almost none of
its competition-level (AMC/AIME/IMO) problems. Evolving directly against benchmarks
of this difficulty means most evaluations end in failure and provide little
selection signal. Conversely, a benchmark easy enough to be informative quickly
saturates if it never hardens.

We therefore use a \emph{coevolving benchmark}, illustrated in
Figure~\ref{fig:overview}. To make benchmark replacement auditable, we
organize the candidate problem pool into three difficulty levels. Although our
longer-term goal is to let agents propose increasingly difficult formal
problems themselves, this study takes a bounded first step: the champion updates
the active benchmark under fixed rules, retiring mastered problems and
replacing them either with new problems from the same level or, once that level
has been mastered, with problems from a harder level. The benchmark can
therefore begin in a tractable regime that provides an effective evolutionary
signal while avoiding the saturation of a fixed easy benchmark. As the agent's proof ability
improves, the benchmark self-hardens, so the proof environment evolves
together with the agent rather than remaining a fixed yardstick.

\paragraph{Why Lean keeps the evolution grounded.}
The formal setting is essential: a self-rewriting proof agent cannot be trusted
to report its own mathematical success; Lean, rather than the agent, is therefore
the arbiter. A proof obligation counts as solved only when the generated proof
re-verifies under a trusted Lean snapshot, and benchmark updates are driven by
these Lean-verified records rather than by self-reports. We additionally require
each attempt to emit an inspectable, Lean-grounded \emph{proof context} whose
representation is free to evolve but whose groundedness is enforced. We detail
this representation-free-but-grounded contract, and the re-verification that
makes solves hard to spoof under the trusted-runtime threat model, in
Section~\ref{sec:contract}.

\paragraph{Contributions.}
\begin{itemize}
	\item \textbf{Self-evolving Lean proof agent.}
	We present a code-level self-evolving Lean theorem-proving agent in which
	the proof workflow and tools are mutable, while proof success remains grounded in Lean
	verification.
	
	\item \textbf{Benchmark coevolution.}
	We introduce a champion-driven, self-hardening benchmark that coevolves
	with the proof agent, using mastery-throttled updates and single-anchor
	recalibration to raise difficulty while keeping scores comparable across
	generations.

	\item \textbf{Representation-free proof contexts grounded by Lean.}
	We require each solve attempt to expose a machine-readable proof context
	whose representation may evolve, while solved claims remain grounded by
	trusted Lean verification.
\end{itemize}

\paragraph{Paper organization.}
Section~\ref{sec:related} situates our work among self-evolving agents,
LLM-based provers, and coevolution / automatic curricula.
Section~\ref{sec:method} presents the system: the mutable workspace and
evolution interface, the verifier-grounded proof-context contract, and the
coevolving-benchmark loop with its parent selection.
Section~\ref{sec:results} reports the experimental study --- capability under the
self-hardening benchmark, a fixed-vs-coevolving ablation, workflow/tool
self-modification, and proof-context quality.
Section~\ref{sec:conclusion} discusses the main implications and concludes.
The appendix collects experimental details, the benchmark-update algorithm, a worked
example, and the lineage record of the accepted run.

\section{Related Work}
\label{sec:related}
Our work sits at the intersection of three lines of research --- self-evolving
agents, LLM-based formal theorem proving, and coevolution with automatic
curricula --- which we review in turn.

\paragraph{Self-evolving agents.}
Recent work on self-evolving agents studies how agents can improve by rewriting
their own code or prompts~\cite{godelmachine,stop,promptbreeder,godelagent,sica},
by searching over agent designs and
workflows~\cite{adas,aflow,agentsquare,evoagent,autoagents}, or by evolving
programs against an automatic evaluator~\cite{funsearch,alphaevolve}, rather
than relying on a fixed hand-designed workflow;
see~\cite{selfevolsurvey,saesurvey} for surveys. DGM~\cite{dgm} evolves an
archive of code-level agents, Hyperagents~\cite{hyperagent} distinguish a task
agent from a meta agent that modifies the improvement process itself, and
related group-evolving systems study population-level
self-improvement~\cite{mae,mage}. Our work transfers this line to Lean theorem
proving and further allows the benchmark to coevolve with the proof agent.

\paragraph{LLM agents for formal mathematical proof.}
Recent work couples LLMs with formal provers through proof generation,
sketching, retrieval, compiler feedback, and repair~\cite{gptf,dsp,leandojo,leancopilot}, with benchmarks such as
miniF2F~\cite{minif2f} and PutnamBench~\cite{putnambench}. LEAP~\cite{leap} and
Goedel-Architect~\cite{goedelarchitect} most directly demonstrate the importance of agentic proof workflows: both build a hand-designed, static workflow around a fixed prover — LEAP maintains its proof plan as an AND-OR DAG of intermediate Lean lemmas, while Goedel-Architect generates and iteratively refines a blueprint, a dependency graph of definitions and lemmas — with Goedel-Architect reaching $99.2\%$ on miniF2F-test. Our work instead makes the proof workflow evolvable, while
requiring Lean-grounded proof context without hard-coding a single representation.

\paragraph{Coevolution and automatic curricula.}
Coevolving solvers with environments or curricula is well established outside
formal proof, including POET, MCC, PAIRED, and automatic-curriculum
methods~\cite{poet,mcc,paired,curriculum,zpd,mastery}. Recent LLM-agent work also
explores agent--environment or agent--benchmark
coevolution~\cite{genenv,mae,mage,deepfact}. Red Queen G\"odel Machine
(RQGM)~\cite{rqgm} coevolves agents with their evaluators: it targets settings
where no fixed ground-truth answer may be available, so the evaluator and the
criterion of success must themselves adapt. Formal mathematical proof, by
contrast, has a fixed correctness criterion and a
strict requirement for proof validity, so we keep the evaluator fixed as a
trusted Lean verifier and coevolve benchmark difficulty instead. Thus, RQGM
coevolves the criterion of success, whereas our system keeps the criterion fixed
and coevolves the difficulty distribution over formally verifiable proof tasks.

\section{Method}
\label{sec:method}
This section presents our system: a fixed, trusted runtime wrapping a mutable,
evolvable workspace (Section~\ref{sec:workspace}); a proof-context contract that
keeps the self-rewriting agent inspectable and hard to spoof
(Section~\ref{sec:contract}); and the coevolving benchmark with its parent
selection (Section~\ref{sec:benchmark}, Section~\ref{sec:selection}).

\subsection{Mutable workspace and evolution interface}
\label{sec:workspace}
Our system separates a fixed, trusted runtime from a mutable workspace. The
trusted runtime is responsible for all mechanisms that determine correctness and
reward --- Lean verification, evaluation, benchmark management, and
proof-context validation ---
and stays outside the evolutionary search. The mutable workspace holds the
agent-facing proof workflow together with the meta-level logic that adapts it,
and is the sole object of evolution: across generations, agents rewrite this
workspace through a constrained edit interface while the trusted runtime remains
untouched.

A central design choice is to separate \emph{code validity} from \emph{proof
success}. A mutated agent first passes through a code-level stage --- basic
execution and smoke tests confirming that its rewritten workflow runs and
returns the expected interface. Passing this stage admits the agent to
evaluation but does not by itself mark any theorem as solved: whether a proof obligation is solved is
decided solely by the trusted runtime, which re-verifies the generated proof
under a trusted Lean snapshot (Section~\ref{sec:contract}). However the agent
rewrites its workflow, tools, or proof-producing code, the solved/unsolved
verdict is determined by Lean alone, and benchmark updates are driven by
Lean-verified solve records rather than by anything the agent reports about
itself. Against this backdrop the seed agent is kept deliberately minimal --- a
single proof attempt that is checked once, with no repair loop, search, or proof
decomposition --- so that any later repair behavior, search strategy, or
proof-context structure can be attributed to the self-evolving process rather
than to a hand-designed proof workflow.

\subsection{Enforced, trustworthy proof context: free to evolve, hard to spoof}
\label{sec:contract}

The proof-context contract has two parts. Part~A specifies what each solve attempt must
emit and what minimal grounding requirements the proof context must satisfy.
Part~B describes the trusted re-verification layer that prevents a mutable
agent from faking solved proofs or proof-context certificates. Together, these
two parts make the agent's proof attempts inspectable while keeping all
reported solves and benchmark records grounded in Lean-verified evidence.

\paragraph{Part A: the contract.}
Every solve attempt returns a triple
$\{\texttt{ok},\,\texttt{proof\_body},\,\texttt{proof\_context}\}$.
The \emph{proof context} records the mathematical structure of the attempt,
such as claims, lemmas, subgoals, dependencies, and how they support the final
theorem. It is kept separate from the workflow trace, which records LLM attempts,
repair loops, and tool calls.

The representation of the proof context is free: the agent may use a graph, a
tree, a lemma table, or another structure. The validator enforces only a small
set of grounding requirements. First, a root context item must exist. Second,
the context must contain at least one mathematical claim, subgoal, or lemma
besides the root. Third, dependency edges, if present, must point to known
items. Fourth, proof-attempt and diagnostic records do not count as mathematical
context. Fifth, any item marked as solved must be backed by Lean-verified
evidence. Finally, if the agent provides optional formal fields such as a
\texttt{lean\_statement} and a \texttt{proof\_body}, the trusted runtime can
verify that item independently and return node-level feedback. The principle is
that representation is flexible, but groundedness is mandatory.

\paragraph{Part B: trusted re-verification and spoofing resistance.}
Because the agent rewrites code adjacent to scoring, none of its self-reports
can be trusted. The trusted runtime therefore re-checks both final proofs and
proof-context certificates before they affect any score or benchmark record.

First, the final score ignores all of the mutable code's self-reported status;
the fixed runtime independently re-verifies the task statement together with the
returned proof body under a trusted Lean snapshot. Second,
top-level Lean commands in the returned proof body are rejected, preventing the
agent from substituting an unrelated theorem. Third, certificates are sanitized: a
proof-context item is marked solved only after its associated Lean evidence
passes trusted verification. Fourth, immutable runtime files and verifier
configuration are protected against tampering; any detected modification is
reverted or rejected, and recorded. Fifth, mutable entry points run in isolated
worker processes on a copied workspace, so the trusted parent does not import
or execute mutable modules directly. Sixth, real Lean proofs are required:
\texttt{sorry} and \texttt{admit} are rejected.

\subsection{Coevolving benchmark}
\label{sec:benchmark}
The benchmark is drawn from a candidate pool stratified into difficulty levels
$L_1 < L_2 < L_3$ (defined in Appendix~\ref{sec:setup}), chosen so that the seed
has substantial success on $L_1$ but low success on $L_3$. Between generations, the
champion drives the benchmark update. A single difficulty coefficient $c$
links the two halves of the loop: it both normalizes the selection scores
(Section~\ref{sec:selection}) and records the accumulated change in benchmark
difficulty.

\paragraph{Benchmark update mechanism.}
The full procedure is given in Algorithm~\ref{alg:update} in
Appendix~\ref{sec:update_algorithm}. At the end of each generation, the
benchmark is updated only if the champion meets the update threshold: its raw
solve rate on the current benchmark must exceed $\gamma=0.30$; otherwise the
benchmark is carried over unchanged.

Once the threshold is met, the update proceeds. A task is
eligible for retirement only if it is solved by all agents evaluated in that
generation, and the update retires at most $\kappa$ such tasks.

For each retired task at level $\ell$, the replacement level is determined by
the champion's current solve rate $m_\ell$ on level $\ell$. If
$m_\ell \ge \tau$ with $\tau=0.70$, the retired task is replaced by a task from
the next level $\ell+1$; otherwise, it is replaced laterally by a new task from
the same level $\ell$. Newly inserted tasks are treated as unsolved until they
are evaluated in a later generation, so $m_\ell$ is recomputed after each
replacement. This makes the update self-throttling: a level can graduate only
while the champion's level-specific mastery remains high, and repeated
replacements reduce the measured mastery rate, preventing abrupt benchmark
hardening.

When selecting replacement tasks, the update also prefers domains where the
archive has accumulated more failures, so the new benchmark reflects both level
difficulty and observed population weaknesses.

\paragraph{Single-anchor recalibration and cross-generation comparability.}
Because the benchmark hardens, raw solve rate is not comparable across
generations. We restore comparability with a \emph{single-anchor}
recalibration: the champion is re-evaluated on the new benchmark and the
difficulty coefficient is updated multiplicatively,
\[
c_{t+1} \;=\; c_{t}\cdot
\frac{r_{t}}
     {\max\!\big(r_{t+1},\,\epsilon\big)},
\qquad c_0 = 1,
\]
where $r_t$ is the champion's raw score on $B_t$, and the
\emph{difficulty-normalized} score reported for every agent is $q = r\times c$. By construction the champion's normalized
score is preserved across the recalibration (it is the anchor), so an agent
evaluated on a harder benchmark and one evaluated on an easier benchmark are
placed on a single scale. The per-step ratio is small (typically
$\lesssim 1.4$ for our pool sizes) but \emph{compounds} across generations, so
the coefficient measures how much harder the current benchmark is than the
original baseline; $\epsilon$ guards against degenerate ratios when the
champion's re-evaluated score is very low.

\subsection{Parent selection within a generation}
\label{sec:selection}
Within a generation, parents are sampled from the archive in proportion to a
DGM-style weight that rewards capability and discounts agents that have already produced many children:
\[
q \;=\; r \times c,
\qquad
w \;=\; \max(q,0)\cdot\frac{1}{1+n_c}.
\]
Here $q$ is the difficulty-normalized score, $r$ the raw solve rate, $c$ the
difficulty coefficient (Section~\ref{sec:benchmark}), and $n_c$ the agent's current
child count; parents are drawn with replacement $\propto w$.
Selection uses \emph{within-generation streaming}: each accepted child is
archived as soon as it is evaluated and immediately becomes eligible as a parent
for later attempts in the same generation, so an early strong child can seed its
own siblings. The child count $n_c$ also includes pending, not-yet-evaluated
children, so the same factor discounts agents with outstanding offspring and
damps runaway selection of a single lineage within a generation.

\section{Results}
\label{sec:results}
This section reports held-out capability across evolution generations, the
comparison with a fixed-benchmark baseline, and the evolution of workflow
structure, mutable tools, and proof-context quality.

During evolution, agents are evaluated on a $76$-task active benchmark
stratified into difficulty levels $L_1$--$L_3$: author-constructed single-tactic
warm-up lemmas, miniF2F problems~\cite{minif2f}, and PutnamBench problems~\cite{putnambench}. To make results comparable across generations, we also
evaluate agents on a fixed held-out miniF2F \emph{test} split of $244$ problems.
This test split is disjoint from the curriculum and is never used for training.
Both the coevolving run and the fixed-benchmark baseline complete $15$ active
generations. For the fixed-benchmark baseline, where raw scores are measured on the same
benchmark throughout, we evaluate the two highest-scoring agents on the
held-out test split. For the coevolving run, where the active benchmark changes over
time, we evaluate selected agents across the trajectory. Backend,
hyperparameters, hardware, and runtime details are provided in
Appendix~\ref{sec:setup}.

\subsection{Capability under a self-hardening benchmark}
Because benchmark difficulty changes throughout evolution, we evaluate
high-scoring agents on the fixed held-out test split, which provides a stable
yardstick for identifying the strongest agent. As shown in
Table~\ref{tab:traj}, held-out solve rate rises
from $12.7\%$ for the seed agent to $45.1\%$ at generation~15. The solve rate
does not increase monotonically, reflecting that selection optimizes
active-benchmark score rather than held-out performance. At the same time, the
benchmark difficulty coefficient rises from $1.00$ to $3.17$; see
Table~\ref{tab:difficulty-coeff} in Appendix~\ref{sec:lineage-record} for
details.

\begin{table}[t]
\centering
\caption{Held-out miniF2F test solve rates for evaluated coevolving agents.}
\label{tab:traj}
\resizebox{\linewidth}{!}{%
\begin{tabular}{lccccccccc}
\toprule
 & Gen 0 & Gen 2 & Gen 4 & Gen 5 & Gen 9 & Gen 10 & Gen 13 &
Gen 14 & Gen 15 \\
\midrule
Solve rate & 12.7\% & 38.9\% & 29.9\% & 29.9\% & 11.9\% & 25.4\% & 40.6\% & 25.4\% & \textbf{45.1\%} \\
Agent & seed & c12 & c26 & c39 & c52 & c86 & c99 & c141 & c144 \\
\bottomrule
\end{tabular}
}
\end{table}

\subsection{Fixed vs.\ coevolving benchmark}
The fixed-benchmark baseline reaches lower held-out solve rates than the coevolving
run. Because the fixed benchmark
never changes, raw training scores are directly comparable; we therefore evaluate
the two highest-scoring fixed-benchmark agents on the held-out test split. Their
held-out solve rates are $32.0\%$ and $26.6\%$. For the
coevolving run, raw training scores are measured under changing benchmark
difficulty, so we evaluate $9$ selected agents across the trajectory. By
generation~15, the best coevolving agent reaches a held-out solve rate of $45.1\%$,
higher than either fixed-benchmark agent.

The fixed-benchmark run scores lower because its evolutionary pressure becomes weak after
early mastery. Once an agent solves a stable subset of the fixed $76$ tasks,
later mutations are selected mainly for preserving or slightly improving
performance on the same problems. The benchmark does not retire mastered tasks or
introduce harder ones, so later agents mostly explore local repair and tool
variants without being forced to broaden their capability. Thus, the
fixed-benchmark run keeps changing code, but mostly around the same narrow set of
solutions, which limits held-out generalization.

\subsection{Workflow self-modification}
\label{sec:workflow_behavior}
The high-scoring workflow is repair-centered, not decomposition-centered.
Across the accepted lineage, selected agents mostly preserve a compact
Lean-error repair loop: generate a full proof, run Lean, use the error message
and lemma-name checks, and retry with bounded feedback
(Figure~\ref{fig:evolution-tree} and Table~\ref{tab:lineage-record}). This winning
workflow pattern is cheap, directly tied to final solve rate, and easier for
selection to reward than a blueprint-first decomposition-based workflow.

This does not mean that decomposition-based workflows are absent from the
search. On the contrary, the mutation traces repeatedly reason about proof
blueprints, subgoals, isolated sub-lemma verification, and deeper proof-context
trees. The gap is between \emph{reasoning about} these workflows and
\emph{shipping} them. Several agents judge the change too large or risky and
fall back to label-only nodes, degenerate whole-theorem nodes, or conventional
repair machinery: c12 writes the binder-isolation fix that verified subnodes
would need but improves instead through lemma tooling, and c39 later defers
isolated sub-lemma verification after observing that variables are unbound. This
behavior is rational under the sparse objective: if only the final
Lean-checked theorem is rewarded, a compact proof body plus compiler-feedback
repair is often a better use of budget than constructing and checking
intermediate proof nodes.

We now examine the decomposition-based workflows that actually evolved: three
main branches shown as bold nodes in the
evolution tree (Appendix~\ref{sec:lineage-record},
Figure~\ref{fig:evolution-tree}).
First, the c22 branch makes an early attempt at a decomposition-and-prove
fallback. This branch tries to add lemma indexing and retrieval machinery around
decomposition, but the implementation hits the mutable-process guard, the trusted
runtime's ban on spawning subprocesses. Its
descendant c23 removes the subprocess-based machinery and simplifies the
workflow, but the simplified version collapses to raw $0.132$. This branch
therefore shows an early failure mode: decomposition needs extra tooling, but
that tooling can easily become too brittle or incompatible with the trusted
runtime. Second, the c44 branch evolves genuine node-level verification, but only in a
shallow form. The c44/c47 lineage asks for self-contained sub-lemmas, carries the
ambient binders into each lemma statement, and verifies many subgoals
independently in Lean. This is a real improvement over post-hoc
\texttt{have}-extraction, but it is not yet a full decomposition-and-assembly
workflow: the branch mainly certifies local helper nodes and does not seriously
address how those verified lemmas should be composed back into the final theorem.
Trace-level problems, such as c44's null root certificate, can also invalidate
otherwise meaningful subgoal evidence. Thus c44/c47 improve proof-context
structure, but remain shallow and do not become the winning route. Third, the c26 lineage is the most important later case because it eventually
pushes decomposition into an explicit plan-prove-assemble workflow. Earlier
agents mostly use decomposition after repair, or only as proof-context structure.
By contrast, the later c141, shown in
Appendix~\ref{sec:example} and Listing~\ref{lst:c141flow}, introduces a
subgoal-prover workflow: it first plans auxiliary lemmas, then tries to prove
them, and finally attempts to assemble the main theorem from those lemmas. This
is the closest the run comes to a decomposition-first workflow. It fails not
because it ignores assembly, but because assembly is exactly the hard part:
generated helper lemmas often do not match the final theorem's needed context,
the interfaces between lemma planning, lemma proving, and final assembly create
extra failure points, and the longer prompts increase timeout and truncation
risk. Since selection rewards only final theorem success, this richer structure
is not protected when its final assembly fails.

Repair itself also does not grow without bound. Proof generation runs under a
bounded output budget and a per-task wall-clock cap (Appendix~\ref{sec:setup}).
When repeated repair accumulates long prompts and error histories,
proof-generation calls can be truncated or time out; these errors are visible to the
agent and become pressure to shorten the next workflow. The large-repair variants
show this saturation: c19 expands the loop to ten attempts but falls to raw
$0.368$, and c51 raises monolithic repair to twelve attempts but scores only
$0.263$. Its descendant c52 keeps the repair idea but shifts toward early
\texttt{native\_decide}, shorter prompts, bans on hallucinated lemmas, and
streamlined result construction. Thus evolution preserves verifier-feedback
repair as a workflow principle, but not unbounded repair: very large repair loops
are competed away.

\subsection{Mutable tools}
\label{sec:tool_analysis}
Most evolved tools target hallucinated Lean names. Early agents add simple
\texttt{\#check} probes and Mathlib source search, so invented Lean names can be
rejected before they enter the final proof. Later agents make the same idea more
systematic: they batch name checks through \texttt{verify\_lemma\_list}, pass
verified signatures into the prompt, and keep small caches or namespace-aware
indexes so that names such as \texttt{Nat.gcd\_dvd\_left} or
\texttt{Nat.dvd\_mul\_right} can be confirmed before being used. c96 turns this
into a cached lemma checker with namespace-aware search and a verified-only
fallback, while c144 feeds exact verified signatures and source-grep suggestions
back into the monolithic repair loop. These tools do not create a new high-level
proving strategy by themselves. Their main role is to make repair safer: when
Lean reports an unknown identifier or type error, the workflow can ban the bad
name, suggest a verified alternative, and retry with concrete Lean feedback.

\subsection{Proof context}
Proof contexts become inspectable but rarely grow into deep verified proof graphs.
Proof-context quality is logged but not rewarded (Section~\ref{sec:selection}), so it should
be read as a diagnostic signal rather than a direct optimization target.
Empirically, it stays roughly stable across generations even as the final
solve rate changes. The lineage record explains why: agents often preserve
inspectable proof context, but rarely turn it into verified mathematical
decomposition trees (Table~\ref{tab:lineage-record}).

The record shows that agents in early and middle generations mostly construct
proof context \emph{backward} from the final Lean proof rather than planning a
proof graph in advance. Agents such as c11, c12, c25, and c26 extract \texttt{have} statements
from the generated proof body and attach them as lemma or subgoal nodes beneath a
single \texttt{main-theorem} node. Their mutation traces explicitly discuss
decomposition-based workflows, subgoal verification, and per-node proof
contexts, but the executed nodes are usually not self-contained Lean theorems:
they depend on binders, local hypotheses, and type context from the surrounding
proof. When isolated for node-level verification, these nodes often fail with
unknown identifiers, type mismatches, or unsolved goals. This phase therefore
increases the number of nodes, but the realized structure is mostly a shallow support star
of the form \texttt{have-*} $\to$ \texttt{main-theorem}, not a deep proof tree or
reusable DAG.

The c44/c47 branch is the first meaningful node-level verification attempt. It
partially solves the binder-isolation problem by asking for self-contained
sub-lemmas whose statements carry the ambient variables as explicit binders, and
by attaching Lean \texttt{lean\_statement}/\texttt{proof\_body} evidence to some
proof-context nodes. Later descendants show both the promise and the limit of
this direction. c99 keeps the c47 structure and produces a
\texttt{subgoal\_tree\_with\_formal\_lemmas} context, with some helper nodes passing
formal verification; however, these nodes remain shallow supports for one final
theorem, not a reusable multi-layer graph. c144 reaches a high solve rate, but
its proof plan mainly guides monolithic repair, and many extracted
\texttt{have} nodes either fail isolated verification or lack the full
\texttt{lean\_statement}/\texttt{proof\_body} pair required for node-level
checking. c145 makes the graph label more explicit as a \texttt{subgoal\_dag},
but its nodes are mostly open or label-only supports and the score drops. Thus
later descendants retain the winning repair-centered mechanisms---lemma
checking, bounded repair, and \texttt{native\_decide} fast paths---while proof
context stabilizes as shallow verified or label-only supports for the main
theorem rather than a deep dependency graph.

\section{Discussion}
\label{sec:conclusion}
This paper studies whether a Lean proof agent can improve by rewriting its own
proof workflow, tools, and proof-context representation while remaining grounded
by a fixed verifier. We build such an agent and pair it with a coevolving
benchmark that hardens as the current champion masters easier tasks. The main
claim is not that the verifier changes or becomes stronger: Lean remains the
trusted judge throughout. Rather, the mutable code around Lean learns better
ways to use verifier feedback, check lemma names, repair failed proofs, and
record proof structure.

The experiments show that useful mathematical workflow structure can evolve
under this fixed-verifier setting. The lineage develops verifier-adjacent tools
for preventing hallucinated Lean names, and it eventually evolves genuine
node-level verification for self-contained sub-lemmas. This supports the
feasibility of node-level verification in a self-modifying proof agent: mutable
agents can invent new proof-context formats, while the fixed runtime can still
audit whether claimed nodes are actually Lean-verified. The fixed-vs-coevolving
ablation also shows why benchmark hardening matters. A fixed benchmark makes
raw scores directly comparable, but it gives weaker curriculum pressure once the
agent reaches a plateau. The coevolving benchmark keeps the training signal
closer to the agent's current frontier.

After $15$ generations, held-out miniF2F test solve rate rises from $12.7\%$
for the seed agent to $45.1\%$ for c144 at generation~15. This is a
substantial improvement from code-level self-evolution alone, but it remains
well below the strongest hand-designed proof agents (Goedel-Architect reaches
$99.2\%$ on miniF2F-test, with a different backend). Our lineage audit suggests a
main reason for the gap: the winning workflow is still mostly repair-centered.
High-scoring agents repeatedly generate a complete proof, ask Lean for errors,
and repair the result with bounded feedback. Decomposition-based workflows do
appear, and later agents increasingly reason about subgoals and node-level
verification, but these structures usually remain shallow and fragile when they
must be assembled back into the final theorem.

We therefore view the current result as evidence for a direction rather than a
finished proof assistant. Longer runs, larger populations, or rewards that
directly value verified decomposition may make decomposition-based workflows a
more important evolutionary route. In the present $15$-generation runs, however,
the agent has not yet evolved a deep, durable proof-context graph. Looking further ahead, we see a natural path in pushing this approach into other domains where formal, Lean-verified reasoning is essential—such as quantum computing—where machine-checked proofs of correctness could prove especially valuable.
\section*{Acknowledgments}
YL, ZW, YG and JL are supported in part by the University of Pittsburgh, School of Computing and Information, Department of Computer Science, Pitt Cyber, Pitt Momentum fund, PQI Community Collaboration Awards, John C. Mascaro Faculty Scholar in Sustainability, Switzerland NSF award 2000-1-243053, NSF award 2535915, Thinking Machines Lab and Cisco Research. This research used resources of the Oak Ridge Leadership Computing Facility, which is a DOE Office of Science User Facility supported under Contract DE-AC05-00OR22725.

\appendix
\section{Experimental details and hyperparameters}
\label{sec:setup}
We report a single-run study; quantitative comparisons therefore come without
variance estimates and should be interpreted accordingly.

\paragraph{Backend, verifier, and hardware.}
A single LLM backend --- DeepSeek \texttt{deepseek-v4-pro} over an
OpenAI-compatible API ($1$M-token context) --- drives both the proof workflow and
its self-modification. We use this backend to keep long evolution runs affordable
while retaining enough capability for Lean proof search. Decoding is greedy
(temperature $0$, fixed random seed). LLM calls use a $600$s API timeout, and
the backend treats max-token truncation as an error rather than passing an
incomplete proof to Lean. Each mutable proof workflow is also run in an isolated
worker with a hard wall-clock cap (set by \texttt{PROOF\_HYPERAGENT\_WORKFLOW\_TIMEOUT\_S};
$1200$s throughout the reported runs). Even at temperature $0$
the backend shows non-negligible output instability on boundary problems, so
small differences should be interpreted cautiously. Correctness is checked by
real Lean~4 with Mathlib (\texttt{v4.30.0}; mock Lean is disabled in scored runs). The evolution
orchestration and all Lean compilation and verification run on a single
multi-core CPU host with the Mathlib environment staged on local storage; the
model is served remotely.

\paragraph{Compute cost.}
A single $15$-generation coevolution run evaluates up to three accepted children
per generation on the $76$-task benchmark and performs the mutation and
smoke-testing steps. Wall-clock time scales roughly linearly with the number of
generations, with held-out test evaluation adding separate benchmark jobs.
Because the model is a remote API, this wall-clock is dominated by Lean
compilation, verification, and long proof-generation calls rather than local
compute.

\paragraph{Benchmark and held-out test set.}
The agent is trained on Lean~4 tasks drawn from a candidate pool stratified into
difficulty levels: $L_1$ consists of author-constructed single-tactic warm-up
lemmas, $L_2$ of miniF2F problems (valid split, disjoint from the held-out test
set), and $L_3$ of PutnamBench problems. These levels impose a deliberate
\emph{evolution gradient}: across the candidate pool, the seed's per-level solve
rate falls steeply ---
$\approx 0.52$ on $L_1$, $\approx 0.18$ on $L_2$, and $\approx 0.00$ on $L_3$ ---
so each level is a rung just beyond the one below. The coevolving active benchmark holds $76$
tasks; at $t{=}0$ its level mix is $L_1{:}L_2{:}L_3 = 27{:}46{:}3$, drawn from a
candidate pool of $365$ tasks ($52$ at $L_1$, $57$ at $L_2$, and $256$ at
$L_3$). We compose this initial mix so that the seed's \emph{overall} solve rate
is $\approx 0.37$ ($28/76$) --- challenging enough to leave room for improvement while
still providing an informative evolutionary signal. Generalization is measured on the held-out
miniF2F \emph{test} split ($244$ problems), disjoint from the valid-split
curriculum and never used for selection.

\paragraph{Experimental hyperparameters.}
Both experiments run for $15$ active evolution
generations after the seed, with up to three accepted children per generation,
aborting after three consecutive backend failures and allowing
at most six mutation attempts per generation. Agent identifiers are global
mutation-attempt IDs across the run, including variants rejected before
evaluation, so they need not match the number of accepted children. The
benchmark-update threshold is a raw champion solve rate of $\gamma = 0.30$; the
graduation threshold is $\tau = 0.70$ and the curriculum step $\sigma = 1$; each
generation retires up to $\kappa = 6$ mastered tasks (the fixed-benchmark
baseline uses $\kappa = 0$), replaces them with domain-matched tasks, and halts
replacement when no eligible task remains in the weak domains. Recalibration
uses $\epsilon = 0.1$ in the multiplicative update
$c_{t+1} = c_t \cdot r_t / \max(r_{t+1}, \epsilon)$, and time and token
penalties are set to $0$.

\section{Benchmark-update algorithm}
\label{sec:update_algorithm}
Algorithm~\ref{alg:update} states the full mastery-throttled benchmark update
whose core mechanisms --- the champion check, weakness-driven selection,
mastery-throttled graduation, self-throttling, and single-anchor recalibration ---
are described in Section~\ref{sec:benchmark}.

\begin{algorithm}[H]
\caption{Mastery-throttled benchmark update (generation $t \to t{+}1$).}
\label{alg:update}
\begin{algorithmic}[1]
\Require active benchmark $B_t$ with curriculum levels; the champion and its
  per-task solve results; candidate pool $P$; update threshold $\gamma$;
  mastery threshold $\tau$; replacement budget $\kappa$; step $\sigma$
\Ensure next benchmark $B_{t+1}$ and difficulty coefficient $c_{t+1}$
\If{the champion's raw solve rate is below $\gamma$}
    \State \Return $B_t$ unchanged \Comment{champion check}
\EndIf
\State $D \gets$ domains most failed across the whole archive \Comment{population weaknesses}
\State $R \gets$ up to $\kappa$ tasks of $B_t$ solved by all agents this generation, lowest level first
\For{each level $\ell$}
    \State $\mathrm{tot}[\ell],\ \mathrm{sol}[\ell] \gets$ champion's total / solved counts on level $\ell$
\EndFor
\State $A \gets \varnothing$
\For{each retired task $q$ at level $\ell$ in $R$}
    \State $m_\ell \gets \mathrm{sol}[\ell]/\mathrm{tot}[\ell]$ \Comment{recomputed after every swap}
    \If{$m_\ell \ge \tau$} \State $\ell^\star \gets \ell + \sigma$ \Comment{graduate}
    \Else \State $\ell^\star \gets \ell$ \Comment{lateral refresh}
    \EndIf
    \State $a \gets$ unused task from $P$ at level $\ell^\star$, preferring domains $D$
    \If{$a = \textsc{nil}$ \textbf{and} $\ell^\star \neq \ell$}
        \State $a \gets$ unused task from $P$ at level $\ell$ \Comment{fallback to lateral}
    \EndIf
    \If{$a = \textsc{nil}$} \State skip this swap
    \Else
        \State $A \gets A \cup \{a\}$; mark $q$ for removal
        \State $\mathrm{tot}[\ell] \gets \mathrm{tot}[\ell]-1$;\quad \textbf{if} the champion solved $q$ \textbf{then} $\mathrm{sol}[\ell] \gets \mathrm{sol}[\ell]-1$ \Comment{self-throttle}
        \State $\mathrm{tot}[\ell^\star] \gets \mathrm{tot}[\ell^\star]+1$ \Comment{added task assumed unsolved}
    \EndIf
\EndFor
\State $B_{t+1} \gets (B_t \setminus R) \cup A$
\State $r_{t+1} \gets$ raw score of the champion re-evaluated on $B_{t+1}$ \Comment{single anchor}
\State $c_{t+1} \gets c_t \cdot r_t/\max(r_{t+1},\,\epsilon)$ \Comment{compounds with an epsilon floor}
\State \Return $B_{t+1},\ c_{t+1}$
\end{algorithmic}
\end{algorithm}

\section{A worked example: an evolved subgoal-prover workflow (agent c141)}
\label{sec:example}
As a worked example we show agent c141, the late-stage child whose evolved
workflow makes the most explicit attempt at a standalone
subgoal-prover workflow. Its meta agent added a \texttt{subgoal\_prover.py}
module: after quick tactics and a short monolithic repair loop, the workflow asks
the model for a lemma plan, proves the proposed lemmas one by one, and then tries
to assemble the final theorem from those lemmas (Listing~\ref{lst:c141flow}).
This is the clearest instance in the run of the structured, node-bearing proof
context that the proof-context contract is meant to make inspectable.

\begin{lstlisting}[caption={Core control flow of c141's evolved
\texttt{solve\_task}: quick tactics, a short monolithic repair loop, then an
explicit subgoal-prover workflow with lemma planning, lemma proofs, and final
assembly.},captionpos=b,label=lst:c141flow]
def solve_task(task, policy, runtime):
    body = _try_quick_tactics(task, runtime)      # 1. fast auto-tactics
    if body is not None:
        return _record(body, "auto_tactic")

    last_solution, ok, attempts, _ = _try_monolithic(
        task, policy, runtime, start_time, tool_text, verified_lemmas)
    if ok:                                      # 2. monolithic repair first
        proof_body = last_solution
        strategy = "monolithic"
    else:
        sub = _try_subgoal(task, policy, runtime, start_time,
                           tool_text, verified_lemmas)
        if sub[0]:                              # 3. explicit subgoal workflow
            proof_body, _, proof_nodes, edges = sub
            strategy = "subgoal"
        else:
            proof_body, ok = _try_safe_mode(
                task, policy, runtime, start_time, verified_lemmas)
            strategy = "safe_fallback" if ok else "monolithic_fallback"
\end{lstlisting}

The subgoal prompt is more ambitious than a \texttt{have}-block parser: it asks
for independent lemma statements with explicit binders, then separately proves
and assembles them (Listing~\ref{lst:c141prompt}).

\begin{lstlisting}[caption={c141's subgoal-planning prompt (abridged): propose
self-contained helper lemmas before proving and assembling them.},
captionpos=b,label=lst:c141prompt]
"Given a theorem, break it down into auxiliary lemmas. "
"Each lemma should be a self-contained statement with all variables explicitly "
"quantified. Do NOT use the theorem's context implicitly. "
"Output up to 5 lemmas as JSON objects with keys: name, statement, explanation."
\end{lstlisting}

Two evolved design choices explain why this more explicit decomposition-based
workflow was nonetheless \emph{competed away}. First, decomposition still runs
after the monolithic route has failed, so it is asked to recover exactly the
hardest cases. Second, the separate lemma-plan, lemma-proving, and assembly
interfaces add new places where the proof can fail: JSON parsing,
non-self-contained lemma statements, and helper lemmas that cannot be introduced
into the final Lean environment, as well as longer prompts and a higher risk of
timeout. The net effect is a structurally richer
workflow whose trace is more informative, but whose solve rate remains below the
repair-centered champions (cf.\ the lineage record in
Appendix~\ref{sec:lineage-record}).

\section{Evolution record of the accepted lineage}
\label{sec:lineage-record}
\begin{figure}[H]
\centering
\includegraphics[width=\textwidth,height=.78\textheight,keepaspectratio]{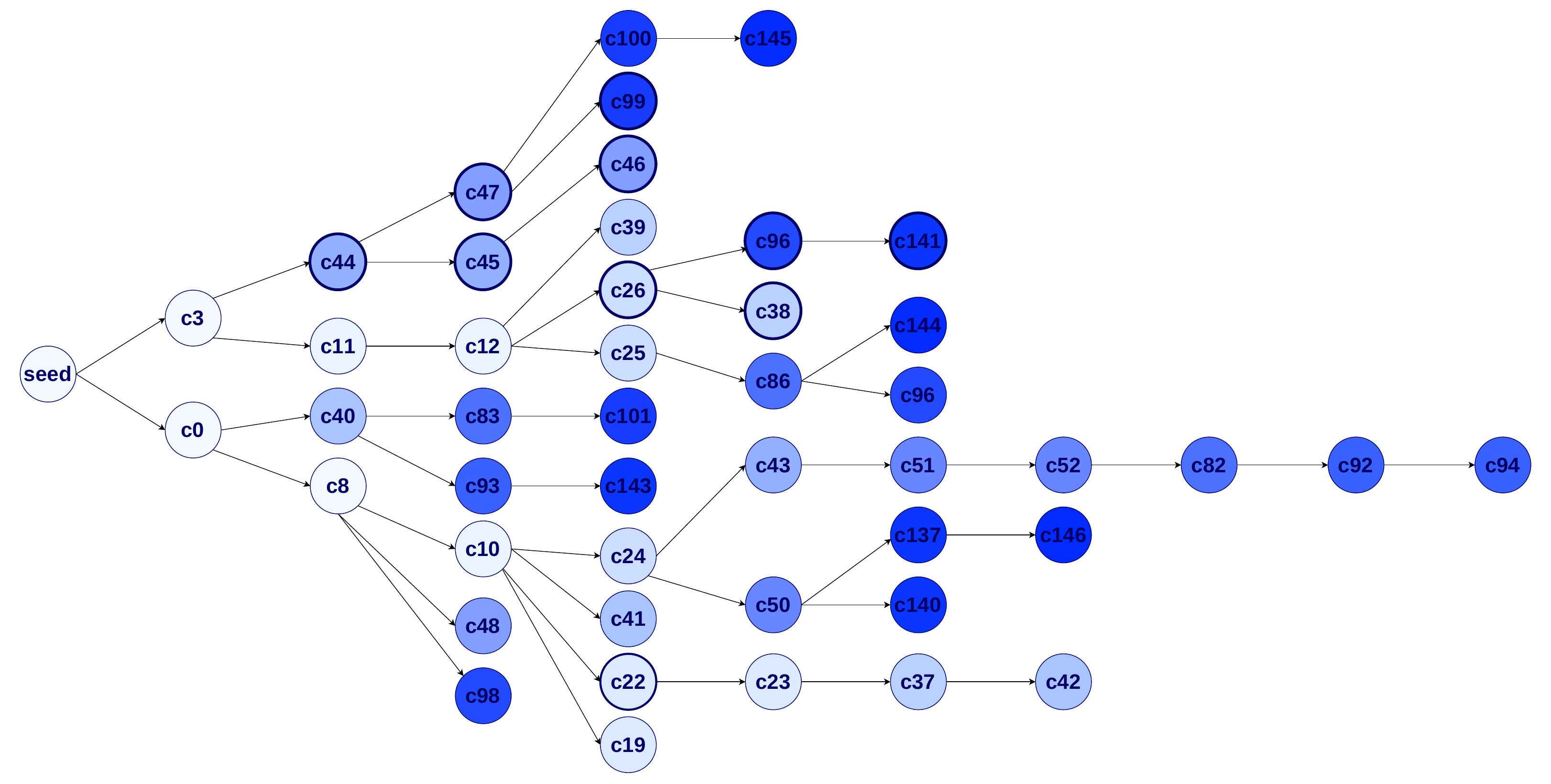}
\caption{Evolution tree for the accepted lineage. Bold nodes mark agents with
decomposition-based workflows. Darker node colors indicate agents evolved in
later generations.}
\label{fig:evolution-tree}
\end{figure}

Section~\ref{sec:example} showed one evolved agent in detail; here we record the
accepted lineage as an evolutionary history, with representative rows for early
generations and, for active generations~10--15, all evaluated accepted children
available at writing time. The goal is to show which
workflow, tool, node-level verification, and proof-context changes were proposed,
which ones actually ran, and which ones survived selection
(Figure~\ref{fig:evolution-tree} and Table~\ref{tab:lineage-record}). The main
lineage-level pattern is that compact repair mechanisms survive, while richer
decomposition-based workflows appear repeatedly but are usually pruned. During
most of the observed run, agents discuss decomposition-based workflows
and per-node verification yet defer them: the sparse pass/fail reward
(Section~\ref{sec:selection}) assigns no gradient to proof structure, so an
agent that maximizes solve rate often skips it. The highest raw-score agent, c12, wrote out
the exact binder fix that node-level verification requires and then skipped it,
improving instead through lemma tooling; a later champion (c39) reasoned that an
isolated sub-lemma ``cannot be verified'' because its variables are unbound and
left it ``for now.''

\begin{table}[H]
\centering
\caption{Difficulty coefficient of the coevolving active benchmark. The
coefficient is used for cross-generation normalization and is not a held-out
solve rate.}
\label{tab:difficulty-coeff}
\resizebox{\linewidth}{!}{%
\begin{tabular}{lcccccccccccccccc}
\toprule
 & Gen 0 & Gen 1 & Gen 2 & Gen 3 & Gen 4 & Gen 5 & Gen 6 & Gen 7 &
Gen 8 & Gen 9 & Gen 10 & Gen 11 & Gen 12 & Gen 13 & Gen 14 & Gen 15 \\
\midrule
Difficulty coeff. & 1.00 & 1.00 & 1.11 & 1.18 & 1.27 & 1.40 & 1.44 &
1.59 & 1.76 & 2.07 & 2.47 & 2.75 & 2.75 & 3.17 & 3.17 & 3.17 \\
\bottomrule
\end{tabular}
}
\end{table}

\begingroup
\scriptsize
\setlength{\LTcapwidth}{\linewidth}
\setlength{\tabcolsep}{2pt}
\renewcommand{\arraystretch}{1.12}
\begin{longtable}{@{}>{\raggedright\arraybackslash}p{1.45cm} *{8}{>{\raggedright\arraybackslash}p{1.67cm}}@{}}
\caption{Four-dimension evolution record of the accepted lineage (single
coevolving run). \emph{Intent}: proposed in the agent's mutation trace;
\emph{Outcome}: shipped, evaluated, and retained or lost under selection.
Raw = solve rate on that generation's benchmark, whose difficulty coefficient
rises $1.00\!\to\!3.17$ by active generation~15, so raw scores are not directly
comparable across generations. \colorbox{gaphi}{Shaded} cells mark proposed
changes that were not realized, not retained, or net-negative; un-shaded
\textsc{done} cells in the
node-level verification column (c44, c47) mark genuinely Lean-certified sub-lemmas.
The workflow columns are restricted to the control-flow choice between repair
and decomposition-based workflows; Lean-name utilities are separated under
mutable tools. \emph{Trace invalid} means the emitted proof context failed the
validator's grounding checks. The seed itself is not listed; c0 is the first
accepted child.
$\star$ generation champion, $\star\star$ highest raw-score agent. Earlier rows
are representative; for active generations~10--15, all evaluated accepted
children available at writing time are shown.}
\label{tab:lineage-record}\\
\toprule
& \multicolumn{2}{c}{\textbf{1. Workflow}}
& \multicolumn{2}{c}{\textbf{2. Mutable tools}}
& \multicolumn{2}{c}{\textbf{3. Node-level verif.}}
& \multicolumn{2}{c}{\textbf{4. Proof-context depth}} \\
\cmidrule(lr){2-3}\cmidrule(lr){4-5}\cmidrule(lr){6-7}\cmidrule(lr){8-9}
Agent (raw) & Intent & Outcome & Intent & Outcome & Intent & Outcome & Intent & Outcome \\
\midrule
\endfirsthead
\caption[]{Four-dimension evolution record of the accepted lineage
(continued).}\\
\toprule
& \multicolumn{2}{c}{\textbf{1. Workflow}}
& \multicolumn{2}{c}{\textbf{2. Mutable tools}}
& \multicolumn{2}{c}{\textbf{3. Node-level verif.}}
& \multicolumn{2}{c}{\textbf{4. Proof-context depth}} \\
\cmidrule(lr){2-3}\cmidrule(lr){4-5}\cmidrule(lr){6-7}\cmidrule(lr){8-9}
Agent (raw) & Intent & Outcome & Intent & Outcome & Intent & Outcome & Intent & Outcome \\
\midrule
\endhead
\midrule
\multicolumn{9}{r}{\footnotesize Continued on next page}\\
\endfoot
\bottomrule
\endlastfoot
c0 ($.368$) & monolithic repair & \textsc{done}: simple retry loop & lemma-name probe & \textsc{part.}: inlined \texttt{\#check} & -- & -- & -- & -- (flat, 1 node) \\
c3 ($.526$) & error-feedback repair & \textsc{done}: repair stays monolithic & lemma existence helper & \textsc{done}: \texttt{\#check} + source search & -- & -- & -- & -- (0 math nodes) \\
c8\,$\star$ ($.553$) & stronger repair loop & \textsc{done}: bounded repair, no decomp. & verified signatures & \textsc{part.}: mostly prompt-side & annotate / decompose into nodes & \gapc{\textsc{n/i}: text-only labels (``unverified'')} & structured context & \textsc{part.}: mild (2 math nodes) \\
\addlinespace[1pt]
c10 ($.447$) & repair hardening & \textsc{part.}: repair policy regresses & decl cache / FD guard & \textsc{part.}: inlined, no durable tool API & whole theorem as one node & \textsc{part.}: degenerate (node $\equiv$ proof) & -- & -- (flat) \\
c11 ($.487$) & repair + shallow decomp. & \textsc{done}: repair first; \texttt{have} nodes post hoc & batch lemma verification & \textsc{done}: edits \texttt{tools.py} & build + isolate-verify \texttt{have} sub-nodes & \gapc{\textsc{n/i}: 7 nodes built, binder-isolation fails} & deep subgoal tree & \textsc{part.}: richest (7 nodes) but unverified \\
c12\,$\star\star$ ($.618$) & compact repair + decomp. idea & \textsc{done}: repair wins; decomp. skipped & verified lemma catalogue & \textsc{part.}: prompt-side, no new tool API & wrote binder fix + planned decomposition verification & \gapc{\textsc{n/i}: skipped; 0 verified sub-nodes} & subgoal tree & \textsc{part.}: $\sim$5 nodes but tooling steps, not math \\
\addlinespace[1pt]
c19 ($.368$) & large repair budget & \gapc{\textsc{n/i}: 10 attempts hurt stability} & expanded lemma cache & \gapc{\textsc{n/i}: no durable advantage} & whole-theorem node & \textsc{part.}: degenerate & -- & -- (flat) \\
c22\,$^{\dagger}$ ($.368$) & decomp. fallback & \gapc{\textsc{n/i}: crash-prone, not selected} & \texttt{grep} index / lookup & \gapc{\textsc{n/i}: guard blocks subprocess; tool dropped} & isolate-verify each \texttt{have} subgoal & \gapc{\textsc{n/i}: deferred (``too risky'')} & LLM sub-lemma skeleton then assemble & \gapc{\textsc{n/i}: skipped (``large change'')} \\
\addlinespace[1pt]
c26\,$\star$ ($.566$) & repair $\to$ decomp. fallback & \textsc{done}: 8 repair attempts, then 4 decomp. & expanded verified-lemma tools & \textsc{done}: prefetch + source search & per-subgoal verify (first real attempt) & \textsc{part.}: runs but hangs on hard tasks, falls back & subgoal tree & \textsc{part.}: built, often monolithic fallback \\
c39\,$\star$ ($.487$) & repair with deferred decomp. & \textsc{done}: repair remains main path & lemma \texttt{\#check} / resolve & \textsc{done}: edits \texttt{tools.py} & isolate-verify but ``vars unbound'' & \gapc{\textsc{n/i}: deferred; \texttt{have}-parsed labels only} & richer context & \textsc{part.}: label-only nodes \\
c41\,$\star$ ($.421$) & monolithic repair & \textsc{done}: no decomp. route & \ttsplit{lemma\_}{exists}\par / cache & \gapc{\textsc{n/i}: mostly inlined; \texttt{tools.py} untouched} & subgoal nodes + per-node verify & \gapc{\textsc{n/i}: cosmetic flag only} & subgoal tree & -- (shallow, 1 claim) \\
c44\,$\star$ ($.408$) & decomp. workflow & \textsc{done}: sublemma-first workflow & cached \texttt{\#check} existence & \textsc{done}: edits \texttt{tools.py} & \textbf{verify self-contained subgoals} & \textbf{\textsc{done}: $141$ certified} (root still null) & real decomposition tree & \textsc{done}: verified shallow graph \\
c47\,$\star$ ($.434$) & extend c44 decomp. & \textsc{done}: two-tier decomp. workflow & verify searched lemmas & \textsc{done}: edits \texttt{tools.py} & verify each subgoal & \textbf{\textsc{done}: $150$ certified}; rest label-only & richest tree (avg $17$) & \textsc{part.}: $173$ verified, $958$ label-only \\
c52\,$\star$ ($.329$) & back to monolithic repair & \textsc{done}: shorter repair; decomp. inactive & complete lemma index & \textsc{done}: edits \texttt{tools.py} & (not selected) & \gapc{\textsc{n/i}: inherited decomposition dead, $0/76$ fires} & -- & -- (flat, 2 nodes) \\
\addlinespace[1pt]
c82 ($.145$) & repair c52 decomp. path & \gapc{\textsc{n/i}: over-constrained; 11/76} & unknown-lemma rejection + short prompts & \textsc{done}: stricter checks, \texttt{force\_short} & inherited subgoal checks & \gapc{\textsc{n/i}: trace valid but low solve rate} & decomp. supports & \textsc{part.}: high score, shallow graph \\
c83 ($.224$) & verify-and-repair loop & \textsc{done}: simple Lean feedback loop & \ttsplit{lemma\_}{exists} + prompt hints & \textsc{part.}: backend errors remain & -- & -- & repair context & \textsc{part.}: shallow; weaker than parent \\
c86\,$\star$ ($.382$) & repair hardening & \textsc{done}: lemma pre-check + bounded repair; no decomp. route & \ttsplit{lemma\_}{checker} / symbol grep & \textsc{done}: new helper, safe-lemma cleanup & node feedback & \gapc{\textsc{n/i}: trace invalid; no durable subgoal verification} & shallow supports & \textsc{part.}: many events, not a deep graph \\
\addlinespace[1pt]
c92 ($.040$) & very large repair budget & \gapc{\textsc{n/i}: 20--25 attempts collapse to 3/76} & signature retrieval & \gapc{\textsc{n/i}: prompt bloat dominates} & proof-node checks & \gapc{\textsc{n/i}: invalid trace} & shallow supports & \gapc{\textsc{n/i}: mostly failures} \\
c93\,$\star$ ($.211$) & verify-and-repair loop & \textsc{done}: simple Lean-feedback loop, low score & def/abbrev index + repair history & \textsc{done}: cached Lean-name checks & -- & -- & repair context & \textsc{part.}: inspectable but shallow \\
c94 ($.066$) & simplify after c92 & \textsc{done}: fast monolithic workflow, still weak & source grep idea & \textsc{part.}: no recovery & -- & -- & flat context & \gapc{\textsc{n/i}: only 5/76} \\
\addlinespace[1pt]
c96 ($.290$) & safe-mode repair & \textsc{done}: verified-only fallback after c95 crash & lemma-checker cache & \textsc{done}: broader namespace search & inherited subgoal parser & \textsc{part.}: parser fixed, not dominant & shallow supports & \textsc{part.}: no new deep graph \\
c97\,$\star$ ($.303$) & decomp.-first workflow & \textsc{done}: plan subgoals, prove, assemble; fallback repair & verified catalogue & \textsc{part.}: safer prompt + lemma filtering & verify each helper lemma & \textbf{\textsc{done}: node-level feedback}; compose fragile & subgoal tree & \textsc{part.}: real but shallow, no breakthrough \\
c98 ($.224$) & multi-sample repair & \textsc{done}: more attempts + namespace scan, weak score & FD guard + lemma ranking & \textsc{part.}: fixes engineering, not solve rate & per-node check & \textsc{part.}: self-contained nodes attempted & math context & \textsc{part.}: shallow, trace invalid \\
\addlinespace[1pt]
c99\,$\star$ ($.290$) & repair + inherited decomp. & \textsc{done}: c47 branch keeps repair/subgoal attempts & fake-lemma autocorrect & \textsc{done}: name map + prompt pitfalls & inherited subgoal checks & \textsc{part.}: not a new certified graph & shallow supports & \textsc{part.}: stable but not deeper \\
c100 ($.263$) & simplify c47 branch & \textsc{done}: plain retry + early unknown-lemma rejection & source signatures & \textsc{done}: grep exact declarations & inherited checks & \textsc{part.}: no new node route & shallow supports & \textsc{part.}: high context, lower solve \\
c101 ($.224$) & repair c83 branch & \textsc{done}: concise retry + reject bad best proofs & cached lemma index & \textsc{done}: forbidden-name list & -- & \gapc{\textsc{n/i}: 1200s timeout appears} & shallow repair context & \textsc{part.}: no structural gain \\
\addlinespace[1pt]
c137 ($.171$) & anti-halluc. repair & \gapc{\textsc{n/i}: stricter checks lower solve rate} & static lemma scan + bad-name rejection & \textsc{done}: prompt + loop edits & \texttt{have}-node extraction & \gapc{\textsc{n/i}: trace invalid, no isolated node checks} & shallow supports & \gapc{\textsc{n/i}: low solve rate, no depth gain} \\
c140 ($.211$) & repair + lint feedback & \textsc{done}: auto-patch then LLM repair, still weak & safe-lemma list / native guard & \textsc{part.}: prompt-side guard & \texttt{have}-node extraction & \gapc{\textsc{n/i}: trace invalid} & shallow supports & \textsc{part.}: inspectable but not deeper \\
c141 ($.250$) & repair decomp. child & \gapc{\textsc{n/i}: decomp. route remains weak} & lemma checker + subgoal prover & \textsc{part.}: helper tooling integrated & subgoal-level verification & \textsc{part.}: helper checks exist, final assembly weak & subgoal tree & \gapc{\textsc{n/i}: many tasks emit no final proof body} \\
c143\,$\star$ ($.276$) & c93 repair + structured prompt & \textsc{done}: lemma checking plus monolithic repair & verified lemma feedback & \textsc{done}: cached name checks in loop & proof-context nodes & \textsc{part.}: nodes logged, not a certified graph & shallow decomp. hints & \textsc{part.}: modest gain over parent \\
\addlinespace[1pt]
c144\,$\star$ ($.303$) & proof-plan prompt + repair & \textsc{done}: plan guides monolithic repair; best held-out solve rate & lemma checker integration & \textsc{done}: verified-lemma hints + stricter prompt & claimed \texttt{have} nodes & \gapc{\textsc{n/i}: flag set, but no isolated node verification} & shallow supports & \textsc{part.}: best score, still no deep DAG \\
c145 ($.211$) & c100 structured context & \textsc{done}: \texttt{have}-extraction, weak solve rate & safe-lemma set + hallucination filters & \textsc{done}: rejects invented names early & extract \texttt{have} subgoals & \gapc{\textsc{n/i}: no node-verif.} & proof-context DAG & \textsc{part.}: richer labels, no gain in solve rate \\
c146 ($.158$) & cached-example repair & \gapc{\textsc{n/i}: extra machinery lowers solve rate} & lemma cache + examples & \textsc{done}: cache wired into workflow & lemma / error feedback & \gapc{\textsc{n/i}: trace invalid, no node checks} & shallow supports & \gapc{\textsc{n/i}: timeout-prone, only 12/76} \\
\end{longtable}

\vspace{2pt}
{\footnotesize $^{\dagger}$c22's lemma index invoked \texttt{subprocess}, which the
mutable-process guard blocks; the uncaught exception crashed nearly every task,
which is why c22 fell to the seed-level score. Its descendant c23 then removed
the index and the decomposition scaffolding in favor of a simpler robust loop.
}
\endgroup

\bibliographystyle{plain}
\bibliography{refs}

@incollection{godelmachine,
  title     = {G{\"o}del Machines: Fully Self-Referential Optimal Universal Self-Improvers},
  author    = {Schmidhuber, J{\"u}rgen},
  booktitle = {Artificial General Intelligence},
  editor    = {Goertzel, Ben and Pennachin, Cassio},
  series    = {Cognitive Technologies},
  pages     = {199--226},
  year      = {2007},
  publisher = {Springer}
}

@article{dgm,
  title   = {Darwin G{\"o}del Machine: Open-Ended Evolution of Self-Improving Agents},
  author  = {Zhang, Jenny and Hu, Shengran and Lu, Cong and Lange, Robert and Clune, Jeff},
  journal = {arXiv preprint arXiv:2505.22954},
  year    = {2025}
}

@inproceedings{godelagent,
  title     = {G{\"o}del Agent: A Self-Referential Agent Framework for Recursively Self-Improvement},
  author    = {Yin, Xunjian and Wang, Xinyi and Pan, Liangming and Lin, Li and Wan, Xiaojun and Wang, William Yang},
  booktitle = {Proceedings of the 63rd Annual Meeting of the Association for Computational Linguistics (Volume 1: Long Papers)},
  pages     = {27890--27913},
  year      = {2025},
  address   = {Vienna, Austria},
  publisher = {Association for Computational Linguistics},
  doi       = {10.18653/v1/2025.acl-long.1354},
  note      = {arXiv:2410.04444}
}

@article{selfevolsurvey,
  title   = {A Survey on Self-Evolution of Large Language Models},
  author  = {Tao, Zhengwei and Lin, Ting-En and Chen, Xiancai and Li, Hangyu and Wu, Yuchuan and Li, Yongbin and Jin, Zhi and Huang, Fei and Tao, Dacheng and Zhou, Jingren},
  journal = {arXiv preprint arXiv:2404.14387},
  year    = {2024}
}

@inproceedings{autoagents,
  title     = {AutoAgents: A Framework for Automatic Agent Generation},
  author    = {Chen, Guangyao and Dong, Siwei and Shu, Yu and Zhang, Ge and Sesay, Jaward and Karlsson, B{\"o}rje F. and Fu, Jie and Shi, Yemin},
  booktitle = {Proceedings of the Thirty-Third International Joint Conference on Artificial Intelligence (IJCAI)},
  year      = {2024},
  doi       = {10.24963/ijcai.2024/3},
  note      = {arXiv:2309.17288}
}

@inproceedings{stop,
  title     = {Self-Taught Optimizer (STOP): Recursively Self-Improving Code Generation},
  author    = {Zelikman, Eric and Lorch, Eliana and Mackey, Lester and Kalai, Adam Tauman},
  booktitle = {Conference on Language Modeling (COLM)},
  year      = {2024},
  note      = {arXiv:2310.02304}
}

@article{promptbreeder,
  title   = {Promptbreeder: Self-Referential Self-Improvement via Prompt Evolution},
  author  = {Fernando, Chrisantha and Banarse, Dylan and Michalewski, Henryk and Osindero, Simon and Rockt{\"a}schel, Tim},
  journal = {arXiv preprint arXiv:2309.16797},
  year    = {2023}
}

@article{adas,
  title   = {Automated Design of Agentic Systems},
  author  = {Hu, Shengran and Lu, Cong and Clune, Jeff},
  journal = {arXiv preprint arXiv:2408.08435},
  year    = {2024}
}

@inproceedings{aflow,
  title     = {AFlow: Automating Agentic Workflow Generation},
  author    = {Zhang, Jiayi and Xiang, Jinyu and Yu, Zhaoyang and Teng, Fengwei and Chen, Xionghui and Chen, Jiaqi and Zhuge, Mingchen and Cheng, Xin and Hong, Sirui and Wang, Jinlin and Zheng, Bingnan and Liu, Bang and Luo, Yuyu and Wu, Chenglin},
  booktitle = {International Conference on Learning Representations},
  year      = {2025},
  note      = {arXiv:2410.10762}
}

@inproceedings{agentsquare,
  title     = {AgentSquare: Automatic LLM Agent Search in Modular Design Space},
  author    = {Shang, Yu and Li, Yu and Zhao, Keyu and Ma, Likai and Liu, Jiahe and Xu, Fengli and Li, Yong},
  booktitle = {International Conference on Learning Representations},
  year      = {2025},
  note      = {arXiv:2410.06153}
}

@inproceedings{evoagent,
  title     = {EvoAgent: Towards Automatic Multi-Agent Generation via Evolutionary Algorithms},
  author    = {Yuan, Siyu and Song, Kaitao and Chen, Jiangjie and Tan, Xu and Li, Dongsheng and Yang, Deqing},
  booktitle = {Annual Conference of the Nations of the Americas Chapter of the Association for Computational Linguistics (NAACL)},
  year      = {2025},
  note      = {arXiv:2406.14228}
}

@article{sica,
  title   = {A Self-Improving Coding Agent},
  author  = {Robeyns, Maxime and Szummer, Martin and Aitchison, Laurence},
  journal = {arXiv preprint arXiv:2504.15228},
  year    = {2025}
}

@article{funsearch,
  title   = {Mathematical discoveries from program search with large language models},
  author  = {Romera-Paredes, Bernardino and Barekatain, Mohammadamin and Novikov, Alexander and Balog, Matej and Kumar, M. Pawan and Dupont, Emilien and Ruiz, Francisco J. R. and Ellenberg, Jordan S. and Wang, Pengming and Fawzi, Omar and Kohli, Pushmeet and Fawzi, Alhussein},
  journal = {Nature},
  volume  = {625},
  pages   = {468--475},
  year    = {2024},
  doi     = {10.1038/s41586-023-06924-6}
}

@article{alphaevolve,
  title   = {AlphaEvolve: A coding agent for scientific and algorithmic discovery},
  author  = {Novikov, Alexander and V{\~u}, Ng{\^a}n and Eisenberger, Marvin and Dupont, Emilien and Huang, Po-Sen and Wagner, Adam Zsolt and Shirobokov, Sergey and Kozlovskii, Borislav and Ruiz, Francisco J. R. and Mehrabian, Abbas and Kumar, M. Pawan and See, Abigail and Chaudhuri, Swarat and Holland, George and Davies, Alex and Nowozin, Sebastian and Kohli, Pushmeet and Balog, Matej},
  journal = {arXiv preprint arXiv:2506.13131},
  year    = {2025}
}

@article{saesurvey,
  title   = {A Survey of Self-Evolving Agents: What, When, How, and Where to Evolve on the Path to Artificial Super Intelligence},
  author  = {Gao, Huan-ang and Geng, Jiayi and Hua, Wenyue and Hu, Mengkang and Juan, Xinzhe and Liu, Hongzhang and Liu, Shilong and Qiu, Jiahao and Qi, Xuan and Wu, Yiran and Wang, Hongru and Xiao, Han and Zhou, Yuhang and Zhang, Shaokun and Zhang, Jiayi and Xiang, Jinyu and Fang, Yixiong and Zhao, Qiwen and Liu, Dongrui and Ren, Qihan and Qian, Cheng and Wang, Zhenhailong and Hu, Minda and Wang, Huazheng and Wu, Qingyun and Ji, Heng and Wang, Mengdi},
  journal = {Transactions on Machine Learning Research},
  year    = {2026},
  note    = {arXiv:2507.21046}
}

@article{gptf,
  title   = {Generative Language Modeling for Automated Theorem Proving},
  author  = {Polu, Stanislas and Sutskever, Ilya},
  journal = {arXiv preprint arXiv:2009.03393},
  year    = {2020}
}

@inproceedings{dsp,
  title     = {Draft, Sketch, and Prove: Guiding Formal Theorem Provers with Informal Proofs},
  author    = {Jiang, Albert Qiaochu and Welleck, Sean and Zhou, Jin Peng and Li, Wenda and Liu, Jiacheng and Jamnik, Mateja and Lacroix, Timoth{\'e}e and Wu, Yuhuai and Lample, Guillaume},
  booktitle = {International Conference on Learning Representations},
  year      = {2023},
  note      = {arXiv:2210.12283}
}

@inproceedings{leandojo,
  title     = {LeanDojo: Theorem Proving with Retrieval-Augmented Language Models},
  author    = {Yang, Kaiyu and Swope, Aidan M. and Gu, Alex and Chalamala, Rahul and Song, Peiyang and Yu, Shixing and Godil, Saad and Prenger, Ryan and Anandkumar, Anima},
  booktitle = {Advances in Neural Information Processing Systems (NeurIPS) Datasets and Benchmarks},
  year      = {2023},
  note      = {arXiv:2306.15626}
}

@inproceedings{leancopilot,
  title     = {Lean Copilot: Large Language Models as Copilots for Theorem Proving in Lean},
  author    = {Song, Peiyang and Yang, Kaiyu and Anandkumar, Anima},
  booktitle = {International Conference on Neuro-symbolic Systems (NeuS)},
  year      = {2025},
  note      = {arXiv:2404.12534}
}

@inproceedings{putnambench,
  title     = {PutnamBench: Evaluating Neural Theorem-Provers on the Putnam Mathematical Competition},
  author    = {Tsoukalas, George and Lee, Jasper and Jennings, John and Xin, Jimmy and Ding, Michelle and Jennings, Michael and Thakur, Amitayush and Chaudhuri, Swarat},
  booktitle = {Advances in Neural Information Processing Systems (NeurIPS) Datasets and Benchmarks},
  year      = {2024},
  note      = {arXiv:2407.11214}
}

@article{leap,
  title   = {LEAP: Supercharging LLMs for Formal Mathematics with Agentic Frameworks},
  author  = {Kung, Po-Nien and Song, Linfeng and Hwang, Dawsen and Yoon, Jinsung and Li, Chun-Liang and Severini, Simone and Ol{\v{s}}{\'a}k, Mirek and Lockhart, Edward and Le, Quoc V. and Gokturk, Burak and Luong, Thang and Pfister, Tomas and Peng, Nanyun},
  journal = {arXiv preprint arXiv:2606.03303},
  year    = {2026}
}

@article{goedelarchitect,
  title   = {Goedel-Architect: Streamlining Formal Theorem Proving with Blueprint Generation and Refinement},
  author  = {Chung, Jui-Hui and Cai, Ziyang and Li, Zihao and Yin, Qishuo and Agarwal, Rohit and Park, Simon and Porto, Rodrigo and Ri, Narutatsu and Yang, Ziran and Tang, Shange and Dang, Xingyu and Lin, Hongzhou and Wang, Mengdi and Chen, Danqi and Jin, Chi and Fowl, Liam H. and Arora, Sanjeev},
  journal = {arXiv preprint arXiv:2606.06468},
  year    = {2026},
  doi     = {10.48550/arXiv.2606.06468}
}

@inproceedings{minif2f,
  title     = {MiniF2F: A Cross-System Benchmark for Formal Olympiad-Level Mathematics},
  author    = {Zheng, Kunhao and Han, Jesse Michael and Polu, Stanislas},
  booktitle = {International Conference on Learning Representations},
  year      = {2022},
  note      = {arXiv:2109.00110}
}

@article{poet,
  title   = {Paired Open-Ended Trailblazer (POET): Endlessly Generating Increasingly Complex and Diverse Learning Environments and Their Solutions},
  author  = {Wang, Rui and Lehman, Joel and Clune, Jeff and Stanley, Kenneth O.},
  journal = {arXiv preprint arXiv:1901.01753},
  year    = {2019}
}

@inproceedings{mcc,
  title     = {Minimal Criterion Coevolution: A New Approach to Open-Ended Search},
  author    = {Brant, Jonathan C. and Stanley, Kenneth O.},
  booktitle = {Proceedings of the Genetic and Evolutionary Computation Conference},
  pages     = {67--74},
  year      = {2017}
}

@inproceedings{paired,
  title     = {Emergent Complexity and Zero-shot Transfer via Unsupervised Environment Design},
  author    = {Dennis, Michael and Jaques, Natasha and Vinitsky, Eugene and Bayen, Alexandre and Russell, Stuart and Critch, Andrew and Levine, Sergey},
  booktitle = {Advances in Neural Information Processing Systems (NeurIPS)},
  year      = {2020},
  note      = {arXiv:2012.02096}
}

@inproceedings{curriculum,
  title     = {Automatic Curriculum Learning For Deep RL: A Short Survey},
  author    = {Portelas, R{\'e}my and Colas, C{\'e}dric and Weng, Lilian and Hofmann, Katja and Oudeyer, Pierre-Yves},
  booktitle = {International Joint Conference on Artificial Intelligence (IJCAI)},
  year      = {2020},
  note      = {arXiv:2003.04664}
}

@book{zpd,
  title     = {Mind in Society: The Development of Higher Psychological Processes},
  author    = {Vygotsky, Lev S.},
  year      = {1978},
  publisher = {Harvard University Press},
  note      = {Edited by Michael Cole, Vera John-Steiner, Sylvia Scribner, and Ellen Souberman}
}

@article{mastery,
  title   = {Learning for Mastery},
  author  = {Bloom, Benjamin S.},
  journal = {Evaluation Comment},
  volume  = {1},
  number  = {2},
  pages   = {1--12},
  year    = {1968}
}

@article{deepfact,
  title   = {DeepFact: Co-Evolving Benchmarks and Agents for Deep Research Factuality},
  author  = {Huang, Yukun and Ribeiro, Leonardo F. R. and Hardalov, Momchil and Dhingra, Bhuwan and Dreyer, Markus and Saligrama, Venkatesh},
  journal = {arXiv preprint arXiv:2603.05912},
  year    = {2026},
  note    = {Accepted at ACL 2026}
}

@article{genenv,
  title   = {GenEnv: Difficulty-Aligned Co-Evolution Between LLM Agents and Environment Simulators},
  author  = {Guo, Jiacheng and Yang, Ling and Chen, Peter and Xiao, Qixin and Wang, Yinjie and Juan, Xinzhe and Qiu, Jiahao and Shen, Ke and Wang, Mengdi},
  journal = {arXiv preprint arXiv:2512.19682},
  year    = {2025}
}

@article{mae,
  title   = {Multi-Agent Evolve: LLM Self-Improve through Co-evolution},
  author  = {Chen, Yixing and Wang, Yiding and Zhu, Siqi and Yu, Haofei and Feng, Tao and Zhang, Muhan and Patwary, Mostofa and You, Jiaxuan},
  journal = {arXiv preprint arXiv:2510.23595},
  year    = {2025}
}

@article{mage,
  title   = {MAGE: Multi-Agent Self-Evolution with Co-Evolutionary Knowledge Graphs},
  author  = {Yang, Ruiyi and Li, Zechen and Xue, Hao and Razzak, Imran and Salim, Flora D.},
  journal = {arXiv preprint arXiv:2605.10064},
  year    = {2026}
}

@article{rqgm,
  title   = {The Red Queen G{\"o}del Machine: Co-Evolving Agents and Their Evaluators},
  author  = {Iacob, Alex and Jovanovi{\'c}, Andrej and Shen, William F. and Burkhardt, Daniel and Kurmanji, Meghdad and Tastan, Nurbek and Sani, Lorenzo and Venanzi, Niccol{\`o} Alberto Elia and Odonnat, Ambroise and Cao, Zeyu and Marino, Bill and Qiu, Xinchi and Lane, Nicholas D.},
  journal = {arXiv preprint arXiv:2606.26294},
  year    = {2026}
}

@article{hyperagent,
	title   = {Hyperagents},
	author  = {Zhang, Jenny and Zhao, Bingchen and Yang, Wannan and Foerster, Jakob and Clune, Jeff and Jiang, Minqi and Devlin, Sam and Shavrina, Tatiana},
	journal = {arXiv preprint arXiv:2603.19461},
	year    = {2026},
	doi     = {10.48550/arXiv.2603.19461}
}

@article{alphaproof,
	title   = {Olympiad-level Formal Mathematical Reasoning with Reinforcement Learning},
	author  = {Hubert, Thomas and Mehta, Rishi and Sartran, Laurent and Horv{\'a}th, Mikl{\'o}s Z. and {\v{Z}}u{\v{z}}i{\'c}, Goran and Wieser, Eric and Huang, Aja and Schrittwieser, Julian and Schroecker, Yannick and Masoom, Hussain and others},
	journal = {Nature},
	volume  = {651},
	pages   = {607--613},
	year    = {2026},
	doi     = {10.1038/s41586-025-09833-y},
	note    = {Published online 2025}
}

@article{deepseekproverv2,
	title   = {DeepSeek-Prover-V2: Advancing Formal Mathematical Reasoning via Reinforcement Learning for Subgoal Decomposition},
	author  = {Ren, Z. Z. and Shao, Zhihong and Song, Junxiao and Xin, Huajian and Wang, Haocheng and Zhao, Wanjia and Zhang, Liyue and Fu, Zhe and Zhu, Qihao and Yang, Dejian and Wu, Z. F. and Gou, Zhibin and Ma, Shirong and Tang, Hongxuan and Liu, Yuxuan and Gao, Wenjun and Guo, Daya and Ruan, Chong},
	journal = {arXiv preprint arXiv:2504.21801},
	year    = {2025},
	doi     = {10.48550/arXiv.2504.21801}
}

@article{goedelproverv2,
	title   = {Goedel-Prover-V2: Scaling Formal Theorem Proving with Scaffolded Data Synthesis and Self-Correction},
	author  = {Lin, Yong and Tang, Shange and Lyu, Bohan and Yang, Ziran and Chung, Jui-Hui and Zhao, Haoyu and Jiang, Lai and Geng, Yihan and Ge, Jiawei and Sun, Jingruo and Wu, Jiayun and Gesi, Jiri and Lu, Ximing and Acuna, David and Yang, Kaiyu and Lin, Hongzhou and Choi, Yejin and Chen, Danqi and Arora, Sanjeev and Jin, Chi},
	journal = {arXiv preprint arXiv:2508.03613},
	year    = {2025},
	doi     = {10.48550/arXiv.2508.03613}
}

\end{document}